%% file: main.tex
\pdfoutput=1

\documentclass[11pt]{article}

\usepackage{emnlp2021}
\usepackage{amsmath,amssymb}
\usepackage{array}
\usepackage{times}
\usepackage{latexsym}
\usepackage{multirow}
\usepackage{graphicx}
\usepackage{subfigure}
\usepackage{booktabs}
\usepackage[T1]{fontenc}

\usepackage[utf8]{inputenc}

\usepackage{microtype}

\title{Solving Aspect Category Sentiment Analysis as a Text Generation Task}

\author{Jian Liu$^{1}$, Zhiyang Teng$^{2,3}$, Leyang Cui$^{2,3}$, Hanmeng Liu$^{2,3}$, Yue Zhang$^{2,3}$ \\
$^1$School of Computer Science, Fudan University\\
$^2$School of Engineering, Westlake University\\
$^3$Institute of Advanced Technology, Westlake Institute for Advanced Study\\
\texttt{jianliu17@fudan.edu.cn},\\
\texttt{\{tengzhiyang, cuileyang, liuhanmeng, zhangyue\}@westlake.edu.cn}\\}

\begin{document}
\maketitle

\input{abstract}

\input{introduction}

\input{relatedwork}

\input{method}

\input{experiment}

\input{analysis}

\section*{Acknowledgements}
Zhiyang Teng is the corresponding author. We would like to thank the anonymous reviewers for their insightful comments. We gratefully acknowledge funding from the National Natural Science Foundation of China (NSFC No.61976180).
\bibliography{anthology,custom}
\bibliographystyle{acl_natbib}

\input{appendix}

\end{document}

%% file: abstract.tex
\begin{abstract}
Aspect category sentiment analysis has attracted increasing research attention. The dominant methods make use of pre-trained language models by learning effective aspect category-specific representations, and adding specific output layers to its pre-trained representation. We consider a more direct way of making use of pre-trained language models, by casting the ACSA tasks into natural language generation tasks, using natural language sentences to represent the output. Our method allows more direct use of pre-trained knowledge in seq2seq language models by directly following the task setting during pre-training. Experiments on several benchmarks show that our method gives the best reported results, having large advantages in few-shot and zero-shot settings.
\end{abstract}

%% file: introduction.tex
\section{Introduction}

Aspect-based sentiment analysis (ABSA) is a fine-grained sentiment analysis task that includes a number of subtasks, two of which are aspect category sentiment analysis (ACSA) and aspect category detection (ACD). Figure \ref{fig:examples} shows an example, where the input is ``\emph{The restaurant was expensive, but the menu was great}''. ACD detects the aspect categories, such as \emph{price} and \emph{food}, and ACSA predicts the sentiment polarities toward each aspect category. In this work, we focus on these two tasks as well as the joint task that combines both.

Previous studies have investigated various methods that treat ACSA and ACD as classification tasks, learning aspect-specific sentence representations \citep{wang2016attention,ruder2016hierarchical}. Recently, pre-trained language models (PLM) have shown their effectiveness to this end \citep{jiang2019challenge}.
The main idea is to make use of pre-trained models such as BERT \cite{bert} for representing an aspect-specific form of the input (e.g., by concatenating the aspect category to the end of the input sentence (Figure \ref{fig:tradition})), which provides useful semantic features for ACSA and ACD classifiers. Such methods have given highly competitive results \cite{sun2019utilizing,LiYZP20}. 

The above classification models benefit from contextualized representations, which contain knowledge learned by pre-training over large data \cite{lin-etal-2019-open}. However, their use of pre-trained knowledge can be viewed as indirect due to at least two reasons. First, the classification task is performed by using a neural network on top of pre-trained representation, with separate network parameters. Second, the integration of aspect category makes the aspect-specific input representation not exactly a natural language sentence, which differs from the pre-training setting. Intuitively, more pre-trained knowledge could be leveraged by connecting pre-training and ACSA at the {\it task} level, rather than only at the {\it representation} level.

\input{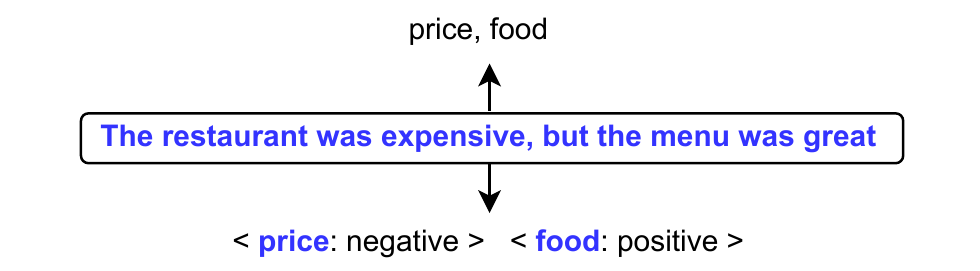}

We investigate the above potentials by casting the sentiment classification tasks into language modelling tasks. In particular, as shown in Figure~\ref{fig:template}, both ACSA and ACD are transformed into sequence-to-sequence (seq2seq) tasks, where the encoder takes the input sentence and the decoder generates a natural language sentence. For ACD, the output follows a template stating whether the specific aspect is discussed (e.g., ``{\it The  $\langle{\tt category\_type}\rangle$ category is discussed}''); for ACSA, the sentiment polarity of a specific aspect is stated (e.g., ``{\it The sentiment polarity of $\langle{\tt given\_category}\rangle$ is  $\langle{\tt polarity\_type}\rangle$}''). The setting corresponds closely to the denoising auto-encoder training scheme of BART \cite{bart}, which we use as the pre-trained model. Compared with classification-based methods, our method does not include more network parameters, and thus can potentially generalize better to new domains \cite{gpt3, gao2020making}. Given a new domain with completely unseen aspect categories and sentiment labels, our method can be applied without changing output layer structure. 
\input{figures/template}

In addition to classification-based methods, we take masked language models (MLM) as a baseline also, for which a natural counterpart of our method is a mask-refilling task. As shown in Figure~\ref{fig:mlm}, different from our method, the output template is concatenated to the input, with the keyword being masked for prediction. This MLM task corresponds closely to BERT \citep{bert} pre-training. In comparison to this MLM method, a generation method can better learn the correlation between the input and output template as two related sequences, which has been demonstrated by the strong performance of BART for abstractive text summarization \cite{bart}.

Experimental results on three standard benchmarks datasets show that both generation and MLM methods outperform classification methods using the same pre-trained language models. Finally, generation methods give stronger performances than MLM methods, outperforming the previous state-of-the-art methods by a large margin. In addition, using the generation method, we show that jointly performing ACSA and ACD leads to better results than the traditional pipeline.
To our knowledge, we are the first to employ a generative pre-trained language model to address an ACSA/ACD problem. We release our code at \url{https://github.com/lgw863/ACSA-generation}.

\input{figures/overview}

%% file: figures/example.tex
\begin{figure}
	\centering
	\setlength{\abovecaptionskip}{0.2cm}
    \setlength{\belowcaptionskip}{-0.5cm}
	\includegraphics[scale=0.8]{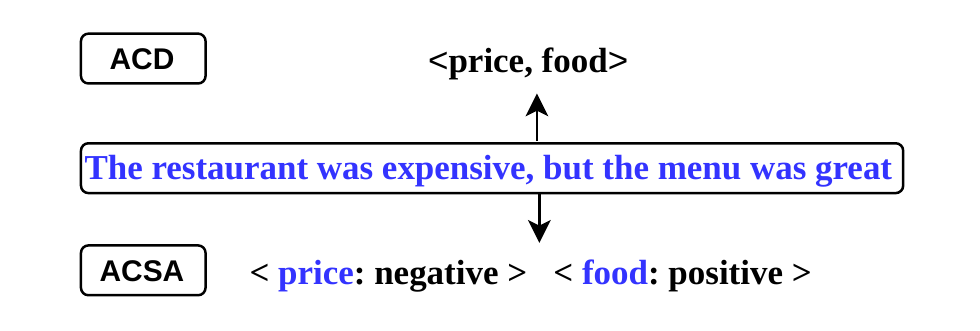}
	\caption{Example of aspect category detection (ACD) and aspect category sentiment analysis (ACSA).} 
	\label{fig:examples}
\end{figure}

%% file: figures/template.tex
\begin{figure}
	\centering
	\setlength{\abovecaptionskip}{0.2cm}
    \setlength{\belowcaptionskip}{-0.4cm}
	\includegraphics[scale=0.8]{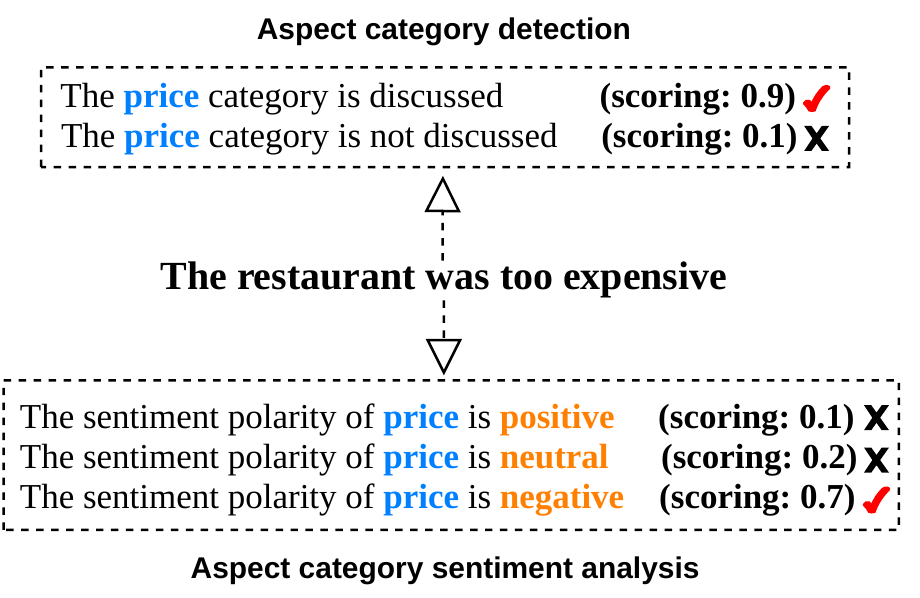}
	\caption{ACSA as a generation task.}
	\label{fig:template}
\end{figure}

%% file: figures/overview.tex
\begin{figure*}[t!]
    \centering
    \setlength{\abovecaptionskip}{0.1cm}
    \setlength{\belowcaptionskip}{-0.5cm}
            \subfigure[BART classification.]{
    \includegraphics[width=0.4\textwidth]{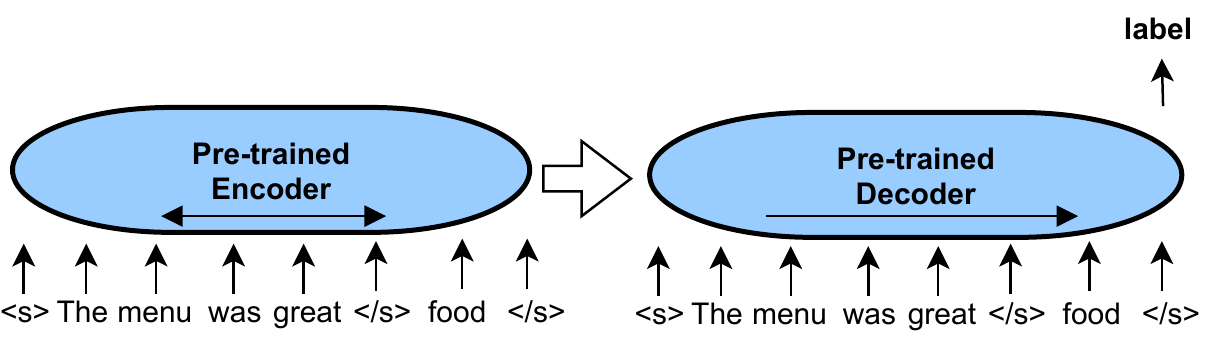}
    \label{fig:tradition}
    }
            \subfigure[Masked language model(MLM).]{
    \includegraphics[width=0.43\textwidth]{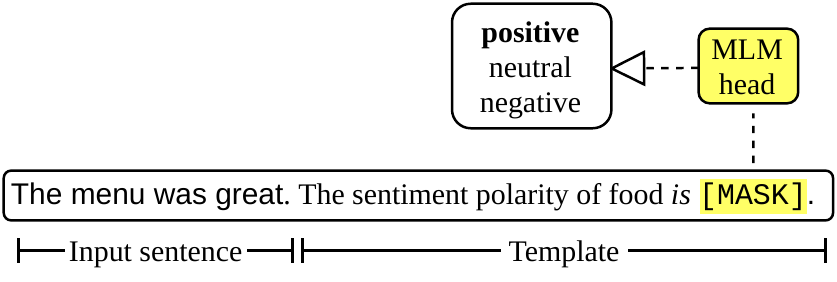}
    \label{fig:mlm}
    }
        \subfigure[BART generation.]{
    \includegraphics[width=0.8\textwidth]{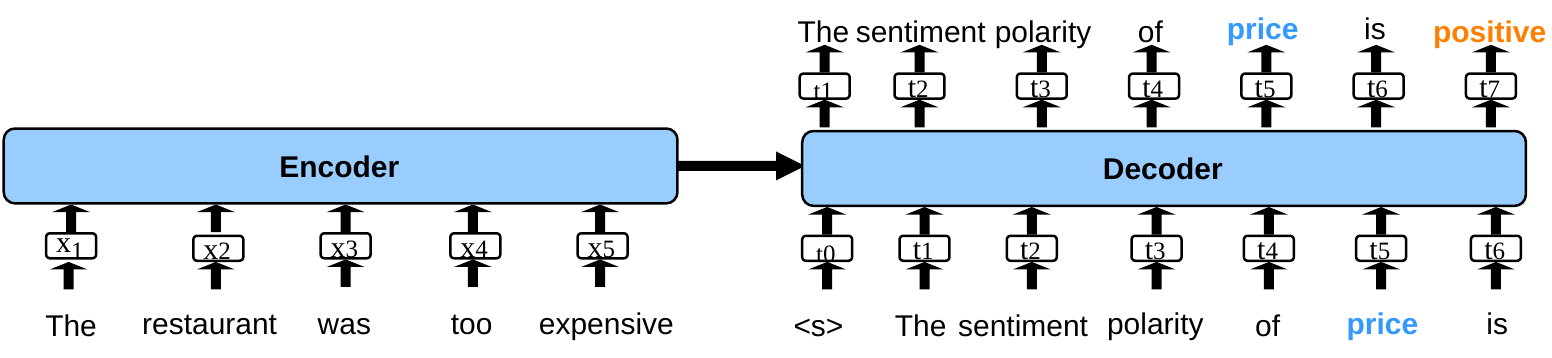}
    \label{fig:temp-train}
    }
    \caption{A comparison of aspect category sentiment analysis methods. \label{fig:overview}}
\end{figure*}

%% file: relatedwork.tex
\section{Related Work}


\textbf{Aspect Category Sentiment Analysis} \citet{wang2016attention} propose an attention-based LSTM network, which can concentrate on different parts of a sentence when different aspect categories are taken as input. \citet{ruder2016hierarchical} model the inter-dependencies of sentences in a text with a hierarchical bidirectional LSTM. \citet{YinSZ17} model the task as a machine comprehension problem by constructing pseudo question-answer pairs. \citet{xue2018aspect} extract sentiment features with CNN and selectively output aspect category related features with gating mechanisms. \citet{xing2019earlier}, \citet{liang2019novel} and \citet{10.1145/3350487} incorporate aspect category information into sentence encoders in the context modeling stage. \citet{sun2019utilizing} construct auxiliary sentences from the aspect categories and convert ACSA to a sentence-pair classification task.
\citet{LiYZP20} predict the sentiment of an aspect category mentioned in a sentence by aggregating the sentiments of the words indicating the aspect category in the sentence. 

Several joint models were proposed to avoid error propagation, which perform ACD and ACSA jointly. \citet{schmitt2018joint} propose two joint models: end-to-end LSTM and end-to-end CNN, which produce all the aspect categories and their corresponding sentiment polarities at once. \citet{hu2019can} propose constrained attention networks (CAN) to constrain the attention weight allocation. \citet{wang2019aspect} propose the aspect-level sentiment capsules model (AS-Capsules), which utilizes the correlation between aspect category and sentiment through shared components. \citet{li-etal-2020-joint-model} propose a novel joint model which contains a shared sentiment prediction layer. 

All the models above are classification methods, which use a separate output network to give the output label. In contrast, we investigate natural language generation methods by directly following the pre-training process of language models.

\textbf{Masked Language Model Methods} There is a line of work using the masked language model (MLM) for natural language understanding tasks. The basic idea is to leverage information from pre-trained models by defining specific sentence prompt in a language modelling task. \citet{gpt3} use prompt for few-shot learning in text classification tasks. \citet{schick2020exploiting} rephrase inputs as cloze questions for text classification. \citet{schick-etal-2020-automatically} and \citet{gao2020making} extend \citet{schick2020exploiting} by automatically generating label words and templates, respectively. \citet{lama} extract relation between entities from BERT by constructing cloze-style templates. We are the first to apply such methods to ACSA, taking it as a baseline. Different from these template-based models, our final model uses BART for text generation, which better models the correlations between the input sentence and the output sentence compared with BERT.

\textbf{Generation Methods}
There has been work casting NLP problems as sequence generation tasks \citep{Vinyals2015GrammarAA,Ma2017DeterministicAF,Stanovsky2018SemanticsAA,journals/jmlr/RaffelSRLNMZLL20}, where the output is a sequence of tokens rather than a natural language sentence. \citet{Daza2018ASM} treat semantic role labelling as a sequence-to-sequence process. \citet{LiYSLYCZL19} solve the entity-relation extraction task as a multi-turn question answering generation method. Our work is similar in casting an NLP task as a generation task. Different from the above methods, our goal is to make the most of pre-trained knowledge in BART for ACSA.


%% file: method.tex
\section{Methods}
Formally for ACD, the input is a sentence $\mathbf{X}=\{x_1,\dots,x_n\}=x_{1:n}$, where $x_i$ denotes the $i$-th word. For ACSA, a set of pre-identified aspect categories are also given. We introduce relevant pre-trained language models in~\ref{sec:lms}, classification methods in Section~\ref{sec:bart}, MLM methods in Section~\ref{sec:mask}, and our generation method in Section~\ref{sec:template-bart}.
\subsection{Pre-trained language Models}
\label{sec:lms}
We take BERT \citep{bert} and BART \citep{bart} as the pre-trained language models. Both are built on the Transformer \citep{transforms} architecture. BERT \citep{bert} is an encoder stack of Transformer for masked text filling, where a model uses the context words to predict masked words. BART \citep{bart} is a denoising auto-encoder seq2seq model pre-training for natural language generation. Its training applies document corruption such as randomly deleting tokens from the input and corrupting text with an arbitrary noising function. BART is trained to reconstruct the original text.


\subsection{The Classification Method}
\label{sec:bart}
We use a multi-layer perceptrons network as the classifier model, which takes a representation vector as input. Both BERT and BART are considered as the encoders.

\paragraph{BERT Classification} BERT adopts “{\it [CLS] input sentence [SEP] given\_category [SEP]}” as input. The final hidden state corresponding to ``[CLS]'' is used as the representation for classification.

\paragraph{BART Classification} BART adopts “{\it $\langle{\tt S}\rangle$ input sentence $\langle{\tt /S}\rangle$ given\_category $\langle{\tt /S}\rangle$}” as input and predicts the sentiment polarity of the sentence towards the given category. The same input is fed into the encoder and decoder (see Figure \ref{fig:tradition}). Formally, suppose that the query category is $a$,  $x_0 = \langle{\tt S}\rangle $, $x_{n+1} = \langle{\tt /S}\rangle$, $x_{n+2} = a$,  $x_{n+3} = \langle{\tt /S}\rangle$, then the input to BART is $x_{0:n+3} = \langle{\tt S}\rangle \ x_1,\dots, x_n \ \langle{\tt /S}\rangle\ a \ \langle{\tt /S}\rangle$. 
The output hidden vectors obtained by the BART encoder (\textsc{Encoder}) and BART decoder (\textsc{Decoder}) are: 
\begin{equation*}
\small 
    \mathbf{h}^{enc} = \textsc{Encoder}(x_{0:n+3}) \\
\end{equation*}
\begin{equation*}
\small
    \mathbf{h}_0 \dots \mathbf{h}_{n+3} = \textsc{Decoder}(\mathbf{h}^{enc}; x_{0:n+3}) \\
\end{equation*}

The output vector $\mathbf{h}_{n+3}$ is then taken as the representation vector for classification. 

\subsection{The MLM Method}
\label{sec:mask}
Masked language models (MLM) \citep{bert} complete a given prompt by filling missing tokens. We refer to the template including a given category and MASK token together as a prompt. For sentiment analysis tasks, \textit{BERT MLM} adopts the input sentence and the prompt as the model input and predicts the sentiment polarity label word towards the given category. For \textit{BART MLM}, the same input is fed into the encoder and decoder, and the highest decoder prediction from label words of the MASK token is the predicted polarity label(see Figure \ref{fig:mlm}). We use the same template in the MLM method and generation method, following the template creation method in section \ref{sec:template}.

\subsection{The Generation Method}
\label{sec:template-bart}
We take both ACSA and ACD as language model ranking problems under a seq2seq framework (see Figure \ref{fig:temp-train}). The target sequence $\mathbf{T}_{a_i,p_k}(\mathbf{T}_{a_i})=\{t_1,\dots,t_m\}$ is a template filled by the given category $a_i$ and the polarity type $p_k$. We first introduce how to create templates in Section~\ref{sec:template}, and then show the inference and training details in Section~\ref{sec:inference} and Section~\ref{sec:training}, respectively.
\subsubsection{Template Creation}
\label{sec:template}
For ACSA, we manually create templates containing one slot for the ${\tt given\_category}$ and another slot for the ${\tt polarity\_type}$ label. We set a category word set $\mathbf{A}=\{a_1,\dots,a_{|C|}\}$, $|C|$ is the category type size (e.g., $a_i$=``{\it price}'') and polarity type word set $\mathbf{P}=\{p_1,\dots,p_{|L|}\}$, $|L|$ is the polarity type size (e.g., $p_k$=``{\it positive}''), and use words to define templates $\mathbf{T}_{a_i,p_k}$ (e.g. ``{\it The sentiment polarity of price is positive}''). The template $\mathbf{T}$ is ``{\it The sentiment polarity of $ \langle a_i \rangle$ is $\langle p_k \rangle$}''. For a given category $a_i$, we can obtain a list of templates $\mathbf{T_{a_i}} = [\mathbf{T}_{a_i,p_1},\dots,\mathbf{T}_{a_i,p_{|L|}}]$.

For ACD, we use $a_i$ to create a sentiment template $\mathbf{T}_{a_i}^+$ for an existing aspect category, and a none-category template $\mathbf{T}_{a_i}^-$. $\mathbf{T}^+$ is ``{\it The $\langle  a_i \rangle$ category is discussed}" and $\mathbf{T}^-$ is  ``{\it The $\langle  a_i \rangle$ category is not discussed}".
\subsubsection{Inference}
\label{sec:inference}
For ACSA, we first enumerate all possible polarities for the given category of the sentence $\mathbf{X}$ and fill them in the prepared templates, and then use the fine-tuned pre-trained generative language model to assign a score for each template $\mathbf{T}_{a_i,p_k} =\{t_1,\ldots,t_m \}$, formulated as:
{\setlength\abovedisplayskip{1pt}
\setlength\belowdisplayskip{1pt}
\begin{equation}
\small
\begin{split}
       f(\mathbf{T}_{a_i,p_k}) = \sum_{c=1}^m \log  P(t_c|t_{1:c-1},\mathbf{X})
\end{split}
\label{score}
\end{equation}}  
\hspace{0.2cm} We calculate a score $f(\mathbf{T}_{a_i,p_k})$ for each possible polarity by employing the pre-trained generative language model (i.e., BART) to score the templates, and then choose the polarity of category $a_i$  with the largest score.


For ACD, we first create templates $\mathbf{T}_{a_i}^+$ and $\mathbf{T}_{a_i}^-$ for all possible categories of the sentence $\mathbf{X}$, and then use the fine-tuned pre-trained generative language model to assign a score for each template $\mathbf{T}_{a_i} =\{t_1,\ldots,t_m \}$, in a similar way as Equation \ref{score}. Also, we decide whether the $a_i$ category is discussed or not in the input sentence according to the higher score between $\mathbf{T}_{a_i}^+$ and $\mathbf{T}_{a_i}^-$.




\subsubsection{Training}
\label{sec:training}
For ACSA, suppose that the polarity type of $a_i$ is $p_k$. We fill the given category $a_i$ and the polarity type $p_k$ into template $\mathbf{T}$ to create a gold target output $\mathbf{T}_{a_i,p_k}$. Similarly for ACD, if the category of $a_i$ is discussed, the gold target $\mathbf{T}_{a_i}^+$ is obtained by filling $a_i$ into $\mathbf{T}^+$, and otherwise is $\mathbf{T}_{a_i}^-$. 

For ACSA, we use all gold polarities in the training set to construct $(\mathbf{X},\mathbf{T})$ pairs. For ACD, we use all gold categories in the training set to construct $(\mathbf{X},\mathbf{T}^+)$ pairs, and additionally create negative samples $(\mathbf{X},\mathbf{T}^-)$ by sampling all none existing categories in the input. Finally, we obtain $\{(\mathbf{X}, \mathbf{T})\} =\{ (\mathbf{X}, \mathbf{T}^{+}) \cup   (\mathbf{X}, \mathbf{T}^-)\}$ 

Given a sequence pair $(\mathbf{X},\mathbf{T})$, we feed the input $\mathbf{X}=x_{1:n}$ to the BART encoder, obtaining hidden representations of the sentence:
{\setlength\abovedisplayskip{2pt}
\setlength\belowdisplayskip{2pt}
\begin{equation}
\small
           \mathbf{h}^{enc} = \textsc{Encoder}(x_{1:n}) 
\end{equation}}
\hspace{0.2cm} At the $c$ th step of the decoder, $\mathbf{h}^{enc}$ and previous output tokens $t_{1:c-1}$ are then as inputs, yielding a representation using attention \cite{transforms}
{\setlength\abovedisplayskip{3pt}
\setlength\belowdisplayskip{3pt}
\begin{equation}
\small
           \mathbf{h}^{dec}_c = \textsc{Decoder}(\mathbf{h}^{enc}, t_{1:c-1}) 
\end{equation}}
\hspace{0.2cm} The conditional probability of the word $t_c$ is defined as:
\begin{equation}
\small
   P(t_c|t_{1:c-1},\mathbf{X}) =  \textsc{softmax}(\mathbf{h}^{dec}_c \mathbf{W}_{lm} + \mathbf{b}_{lm}),
\label{eq:bart_output}
\end{equation}
where $\mathbf{W}_{lm} \in \mathbb{R}^{d_h \times |\mathcal{V}|}$ and $\mathbf{b}_{lm} \in \mathbb{R}^{|\mathcal{V}|}$, $|\mathcal{V}|$ represents the vocab size of pre-trained BART. The cross-entropy between the decoder’s output and the original template is used as the loss function:
{\setlength\abovedisplayskip{2pt}
\setlength\belowdisplayskip{0pt}
\begin{equation}
\small
   \mathcal{L} = - \sum_{c=1}^m \log P(t_c|t_{1,c-1}, \mathbf{X})
\end{equation}}

%% file: experiment.tex
\section{Experiments}

We choose the SemEval-2014 restaurant review (Rest14) \cite{pontiki-etal-2014-semeval}, a variant of Rest14 (Rest14-hard) \cite{xue2018aspect} and the multi-aspect multi-sentiment (MAMS) \cite{jiang2019challenge} datasets for sentence-level sentiment , the TripAdvisor \cite{WangLZ10} and BeerAdvocate \cite{McAuley,Lei} datasets for document-level sentiment. Standard splits of training/development/testing sets are adopted following previous work \citet{tay2018learning}, the details of which are shown in Appendix A. 
We use the pre-trained BERT-base\footnote{\url{https://github.com/google-research/bert}} and BART-base\footnote{\url{https://huggingface.co/facebook/bart-base/tree/main}} models for task fine-tuning. 
We select the fine-tuning learning rate from \{4e-5, 2e-5, and 1e-5\} and batch size from \{8, 16, 24\} for different models. The dropout probability is 0.1. 
The best model configuration is selected according to the highest performance on the development set. The details of settings are shown in Appendix~\ref{sec:appendix}.

\subsection{Baseline Methods}
We compare our generation method with classification and MLM baselines (Figure \ref{fig:overview}) using the same encoder. In particular, \textit{BART generation} (i.e., Figure \ref{fig:temp-train}) is compared with \textit{BART classification} (Figure \ref{fig:tradition}) and \textit{BART MLM} (Figure \ref{fig:mlm}), as well as \textit{BERT classification} and \textit{BERT MLM}. In addition, our method is also compared with other models in the literature as follows.

For sentence-level ACSA, we also compare our method with the following state-of-the-art methods in the literature. (1) non-BERT models: GCAE \citep{xue2018aspect}, As-capsule \citep{wang2019aspect} and CapsNet \citep{jiang2019challenge}; (2) BERT \citep{devlin2019bert} based models: BERT-pair-QA-B \citep{sun2019utilizing}, CapsNet-BERT \citep{jiang2019challenge} and AC-MIMLLN-BERT \citep{LiYZP20}. 


For document-level ACSA, we compare our method with the following methods. (1) non-BERT models: LSTM \citep{TangQL15}, HAN \citep{YangYDHSH16} and MR (machine comprehension pattern) \citep{YinSZ17}; (2) BERT \citep{devlin2019bert} based model: \textit{BERT classification}.

For ACD, we compare our method with the following methods. (1) non-BERT models: XRCE \citep{BrunPR14}, NRC-Canada \citep{KiritchenkoZCM14}; (2) BERT \citep{devlin2019bert} based models: \textit{BERT classification}, BERT-pair-NLI-B \citep{sun2019utilizing}
, CNE-net \citep{DaiPCD20}.

\input{tables/template1}
\input{tables/template2}

\subsection{Development Experiments}
Different templates can be used for expressing the same meaning. For instance, ``{\it The sentiment polarity of $\langle{\tt given\_category}\rangle$ is positive}'' can also be expressed by ``{\it The sentiment is positive for $\langle{\tt given\_category}\rangle$}''. For ACSA, we investigate the impact of manual templates using the MAMS development set. Table \ref{tab:ACSA_temp} shows the impact of different choice of templates. For instance, ``{\it The $\langle{\tt given\_category}\rangle$ category has a  $\langle{\tt polarity\_type}\rangle$ label}'' and ``{\it The sentiment polarity of $\langle{\tt given\_category}\rangle$ is  $\langle{\tt polarity\_type}\rangle$}'' give 82.31\% and 83.78\% accuracy, respectively, indicating that the template has influence on the final performance. This is consistent with finds of \citet{gao2020making} for the few-shot task. Based on the development results, we use the top performing template ``{\it The sentiment polarity of $\langle{\tt given\_category}\rangle$ is $\langle{\tt polarity\_type}\rangle$}'' in our ACSA experiments.

\input{tables/acsa_sentence}
\input{tables/acsa_document}

For ACD, we investigate the impact of templates using the Rest14 development set. Table~\ref{tab:ACD_temp} shows the performance impact of different templates. We use the top performing template ``{\it The  $\langle{\tt category\_type}\rangle$ category is discussed}'' as template $\mathbf{T}^+$ and ``{\it The  $\langle{\tt category\_type}\rangle$ category is not discussed}'' as template $\mathbf{T}^-$ in our ACD experiments.

\subsection{ACSA Experiments}

The results of sentence-level ACSA are shown in Table~\ref{tab:acsa_sentence}. We can see that, first, the performance of \textit{BERT MLM} and \textit{BART MLM} is better than \textit{BERT classification} and \textit{BART classification}, respectively. In particular, \textit{BERT MLM} gives a strong baseline, outperforming all non-BERT and \textit{BERT classification} baselines. This shows that making use of pre-training at the {\it task} level can achieve better results than that at the {\it representation} level. Also, the BART \textit{MLM} and \textit{classification} models perform better than the corresponding BERT models. Second, \textit{BART generation} outperforms all baselines on all three datasets, which indicates that our model can better detect multiple sentiment polarities in one sentence toward different aspect categories. 
Third, \textit{BART generation} performs significantly better than \textit{BART MLM}, giving absolutely 3.89\% stronger accuracy on MAMS, demonstrating the effectiveness of the generation method. This shows the strength of BART pre-training for generating semantically related content, which was also reflected by the strong performance of BART on abstractive summarization \cite{bart}. In contrast, the MLM method concatenates the input and output into one sequence, and thus fails to model their correlation in encoder-decoder pre-training.

The performances of our model on document-level ACSA are shown in Table~\ref{tab:acsa_document}. Compared with LSTM, HAN and MR, \textit{BERT classification} and \textit{BART classification} outperform all baselines, which shows the effectiveness of pre-training. \textit{BERT MLM} and \textit{BART MLM} surpass \textit{BERT classification} and \textit{BART classification}, respectively. 
Our \textit{BART generation} model achieves improvements of 1.15\% and 0.70\% over \textit{BART MLM} on TripAdvisor and BeerAdvocate, respectively, demonstrating that the generation method can more effectively make use of BART for ACSA.
\input{tables/acd_rest14}

\subsection{ACD Experiments}
\input{tables/combination}
\input{figures/few_shot}
Results on the Rest14 ACD subtask are presented in Table \ref{tab:acd_rest14}. Following \citet{PontikiGPPAM14}, we use Micro-F1 for evaluating. Again \textit{BART generation} achieves better results than \textit{BART classification} and \textit{BART MLM}. Our model outperforms all baselines on precision and F-1 score. In particular, a more than 95\% precision score is achieved, which shows that our model can effectively exclude the aspect categories not mentioned in the input.
\input{tables/acd_mams}

We also investigate the performance on the MAMS dataset, which consists of at least two unique aspect categories with different sentiment polarities in each input sentence. Table \ref{tab:acd_mams} shows that \textit{BART generation} outperforms all baselines, indicating better ability of our model to detect multiple aspect categories in one sentence.

\subsection{A Joint Model}
The generation method allows us to build a straightforward joint model by extending the first template in Table \ref{tab:ACSA_temp}, using ``{\it The sentiment polarity of <given\_category> is none}'' as a template for non-existing aspect categories. The results on Rest-14 and MAMS are presented in Table \ref{tab:combination}. We find that joint \textit{BART generation} achieves better results on this task with improvements over pipeline \textit{BART generation}. Joint \textit{BART generation} outperforms all baselines on precision, recall and F-1 score, which shows the advantage of joint learning. 

\input{tables/zero_shot}

\subsection{Few-Shot and Zero-Shot Learning}

We evaluate the model performance on ACSA where only a small amount of labelled data is available for training, simulating the low-resource data scenarios by randomly sampling training instances from a large training set. In particular, we use different numbers of instances for training, randomly sampling a fixed number of instances per category type (10, 20, 50, 100, 200, 500 instances per category type for Rest14 and MAMS). The results are shown in Figure~\ref{tab:few_shot}, where the methods of \textit{BERT classification}, \textit{BART classification} and \textit{BART MLM} are also compared. 

It can be seen that on all the datasets, our model outperforms \textit{BERT classification}, \textit{BART classification} and \textit{BART MLM}, especially when the number of training instances is small. For example, when there are only 10 training instances, our model gives accuracy scores of 82.01\% on Rest14, as compared to 38.57\% by \textit{BERT classification} and 50.16\% by \textit{BART classification}. When the number of instances grows as large as 500, our model gives 2.24\% and 2.65\% better accuracies than \textit{BART MLM} on Rest14 and MAMS, respectively. One possible reason is that our method makes more use of direct sentiment knowledge in the pre-trained language model by directly adopting the original structure of BART mentioned earlier. In contrast, classification methods cannot achieve this due to transferring the sentiment bias indirectly.

The results of our zero-shot learning experiments are in Table \ref{tab:zero_shot}. In all cases, our method outperforms all the baselines. In particular, the model trained on MAMS has a better performance on Rest14 than the reverse zero-shot setting, which proves that the MAMS dataset has a higher challenge.

%% file: tables/template1.tex
\begin{table}[t!]
\small
    \centering
	\setlength{\abovecaptionskip}{0.1cm}
    \setlength{\belowcaptionskip}{-0.4cm}
    \begin{tabular}{c|c}
    \toprule
          ACSA Template $\mathbf{T}$ & Dev accuracy \\
    \midrule
        The sentiment polarity of $\mathbf{a}_{i}$ is $\mathbf{p}_{k}$ & \textbf{83.78}\\
        The sentiment is $\mathbf{p}_{k}$ for $\mathbf{a}_{i}$ & 83.44 \\
        The $\mathbf{a}_{i}$ category has a $\mathbf{p}_{k}$ label  & 82.31\\
    \bottomrule
    \end{tabular}
    \caption{ACSA results using different templates. $\mathbf{a}_{i}$ indicates given category, $\mathbf{p}_{k}$ indicates polarity type.}
    \label{tab:ACSA_temp}
\end{table}

%% file: tables/template2.tex
\newcommand{\tabincell}[2]{\begin{tabular}{@{}#1@{}}#2\end{tabular}}
\begin{table}[t!]
\small
    \centering
	\setlength{\abovecaptionskip}{0.1cm}
    \setlength{\belowcaptionskip}{-0.6cm}
    \begin{tabular}{c|c}
    \toprule
          ACD Template $\mathbf{T}^+$/$\mathbf{T}^-$ & Dev F1 \\
    \midrule
    \tabincell{l}{The $\mathbf{a}_{i}$ category is discussed \\ 
    The $\mathbf{a}_{i}$ category is not discussed} & \textbf{93.13} \\
	     
    \midrule
    \tabincell{c}{The sentence discusses the $\mathbf{a}_{i}$ category \\ 
    The sentence discusses no $\mathbf{a}_{i}$ category} & 92.67 \\
	     
    \midrule
    \tabincell{l}{It is about the $\mathbf{a}_{i}$ category \\ 
    It is not about the $\mathbf{a}_{i}$ category} & 92.44 \\
    \bottomrule
    \end{tabular}
    \caption{ACD results using different templates. $\mathbf{a}_{i}$ indicates category type.}
    \label{tab:ACD_temp}
\end{table}

%% file: tables/acsa_sentence.tex
\begin{table*}[t!]
\small
	\centering
	\setlength{\abovecaptionskip}{0.2cm}
    \setlength{\belowcaptionskip}{-0.4cm}
	\begin{tabular}{p{2cm}|l|l|l|l } 
		\toprule
		Category & Model & Rest14 & Rest14-hard & MAMS \\
		\midrule
		\multirow{4}{2cm}{\centering Classification w/o PLM} & GCAE \citep{xue2018aspect} & 81.336($\pm$0.883) & 54.717($\pm$4.920) & 72.098$\dagger$\\
		& As-capsule \citep{wang2019aspect} & 82.179($\pm$0.414) & 60.755($\pm$2.773) & 75.116($\pm$0.473)\\
		& CapsNet \citep{jiang2019challenge} & 81.172($\pm$0.631) & 53.962($\pm$0.924) & 73.986$\dagger$\\
		& AC-MIMLLN \citep{LiYZP20} & 81.603($\pm$0.715) & 65.283($\pm$2.264) & 76.427($\pm$0.704)\\
		\midrule
		\multirow{6}{2cm}{\centering Classification w PLM} & BERT classification & 87.482($\pm$0.906) & 67.547($\pm$5.894) & 78.292$\dagger$\\
		& BART classification  & 88.289($\pm$0.943) & 68.698($\pm$3.407) & 78.761($\pm$0.752)\\
		& BERT-pair-QA-B \citep{sun2019utilizing} & 87.523($\pm$1.175) & 69.433($\pm$4.368) & 79.134($\pm$0.973)\\
		& CapsNet-BERT \citep{jiang2019challenge} & 86.557($\pm$0.943) & 51.321($\pm$1.412) & 79.461$\dagger$\\
		& AC-MIMLLN-BERT \citep{LiYZP20} & 89.250($\pm$0.720) & 74.717($\pm$3.290) & 81.198($\pm$0.606)\\

		\midrule
		\multirow{2}{2cm}{\centering Masked language model} & BERT MLM & 88.446($\pm$0.825) & 69.021($\pm$2.753) & 79.019($\pm$0.935)\\
		& BART MLM & 88.667($\pm$0.768) & 69.585($\pm$2.529) & 79.243($\pm$0.854)\\
		\midrule
		Generation & BART generation  & \textbf{90.545($\pm$0.315)}$^*$ & \textbf{77.358($\pm$2.160)}$^*$ & \textbf{83.130($\pm$0.478)}$^*$\\
		\bottomrule
	\end{tabular}
	\caption{Results of the sentence-level ACSA in terms of accuracy (\%, mean$\pm$(std)). $\dagger$ refers to \citet{jiang2019challenge}. * means the result is significant at $p < 0.01$ using paired t-test comparing to \textit{BART MLM} and \textit{BART classification}.}
	\label{tab:acsa_sentence}
\end{table*}

%% file: tables/acsa_document.tex
\begin{table}[t!]
\small
	\centering
	\setlength{\abovecaptionskip}{0.2cm}
    \setlength{\belowcaptionskip}{-0.4cm}
	\begin{tabular}{l|c|c}
		\toprule
		Model & TripAdvisor & BeerAdvocate  \\
		\midrule
		LSTM & 44.02  & 34.78    \\
		HAN & 44.68  & 36.03   \\
		MR & 46.56  & 38.06   \\
		\midrule
		BERT classification & 47.03  & 39.85    \\
		BART classification & 47.45  & 40.44    \\
		\midrule
		BERT MLM & 48.03  & 40.58    \\
		BART MLM & 48.36  & 40.72    \\
		\midrule
		BART generation & \textbf{49.51}$^*$  & \textbf{41.42}$^*$   \\
		\bottomrule
	\end{tabular}
	\caption{Results of the document-level ACSA in terms of accuracy (\%). * means the result is significant at $p < 0.01$ using paired t-test comparing to \textit{BART MLM} and \textit{BART classification}.
	\label{tab:acsa_document}
	}
\end{table}

%% file: tables/acd_rest14.tex
\begin{table}[t!]
\small
	\centering
	\setlength{\abovecaptionskip}{0.2cm}
    \setlength{\belowcaptionskip}{-0.4cm}
	\begin{tabular}{l|c|c|c}
		\toprule
		Model & P & R & F1  \\
		\midrule		
		XRCE & 83.23 & 81.37 & 82.29 \\
		NRC-Canada & 91.04 & 86.24 & 88.58 \\
		\midrule
		BERT classification & 92.78 & 89.07 & 90.89 \\
		BERT-pair-NLI-B & 93.57 & \textbf{90.83} & 92.18 \\
		CNE-net & 93.76 & \textbf{90.83} & 92.27 \\
		\midrule
		BART classification & 93.01 & 89.92 & 91.44 \\
		BART MLM & 93.44 & 89.83 & 91.60 \\
		BART generation & \textbf{95.18} & 90.54 & \textbf{92.80} \\
		\bottomrule
	\end{tabular}
	\caption{Rest14 results: Aspect Category Detection. We use the results reported in XRCE \cite{BrunPR14}, NRC-Canada \cite{KiritchenkoZCM14}, BERT-pair-NLI-B \citep{sun2019utilizing} and CNE-net \citep{DaiPCD20}.
	\label{tab:acd_rest14}
	}
\end{table}

%% file: tables/combination.tex
\begin{table*}[t!]
\footnotesize
    \centering
	\setlength{\abovecaptionskip}{0.2cm}
    \setlength{\belowcaptionskip}{-0.4cm}
	\begin{tabular}{l| c| c| c| c| c| c}
		\toprule
		\multirow{2}*{Model} & \multicolumn{3}{c|}{Rest14} &  \multicolumn{3}{c}{MAMS}\\
		\cmidrule{2-7}
		~ & P & R & F1 & P & R & F1\\
		\midrule
		Pipeline BART generation & 82.03 & 76.46 & 79.15 & 77.04 & 71.92 & 74.39\\
		\midrule
		Joint BERT classification & 77.75  & 76.07  & 76.90  & 74.14  & 71.92  & 73.01 \\
		Joint BART classification & 81.92 & 73.59 & 77.53 & 74.59  & 74.13 & 74.36 \\
		Joint BART MLM & 81.88 & 76.73 & 79.22 & 75.32  & 75.07 & 75.19 \\
		Joint BART generation & \textbf{82.76} & \textbf{81.91} & \textbf{82.33} & \textbf{77.18} & \textbf{76.58} & \textbf{76.88}\\
		\bottomrule
	\end{tabular}
	\caption{\label{tab:combination} Performance on combination setting. 
	}
\end{table*}

%% file: figures/few_shot.tex
\begin{figure*}[t!]
    \centering
    \setlength{\abovecaptionskip}{0cm}
    \setlength{\belowcaptionskip}{-0.4cm}
            {
    \includegraphics[width=0.38\textwidth]{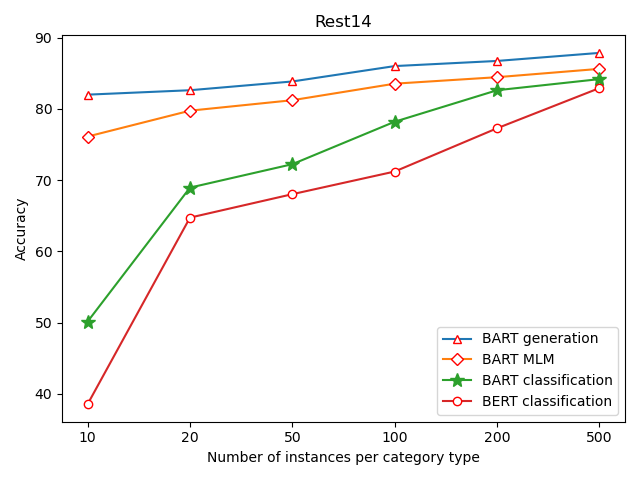}
    \label{fig:few_rest14}
    }
            {
    \includegraphics[width=0.38\textwidth]{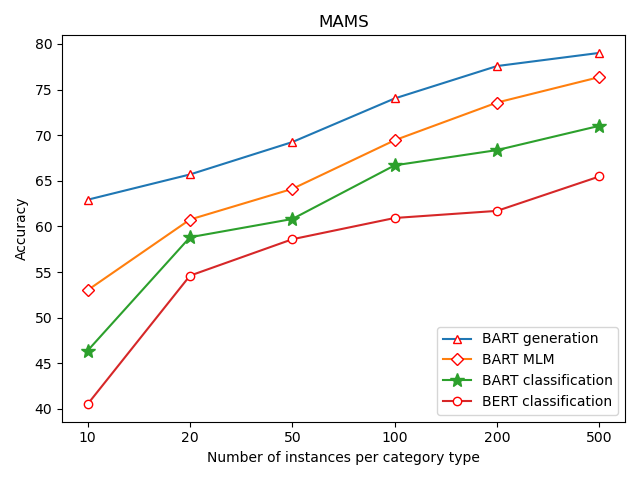}
    \label{fig:few_mams}
    }
    \caption{\label{tab:few_shot}Few-shot ACSA performance on different test sets.}
\end{figure*}


%% file: tables/acd_mams.tex
\begin{table}[t!]
\small
	\centering
	\setlength{\abovecaptionskip}{0.2cm}
    \setlength{\belowcaptionskip}{-0.4cm}
	\begin{tabular}{l|c|c|c}
		\toprule
		Model & P & R & F1  \\
		\midrule
		BERT classification & 90.50  & 86.68  & 88.50  \\
		BART classification & 90.67  & 88.34  & 89.49  \\
		BART MLM & 90.57  & 88.86  & 89.71  \\
		BART generation & \textbf{90.71}  & \textbf{90.16}  & \textbf{90.43}  \\
		\bottomrule
	\end{tabular}
	\caption{MAMS results: Aspect Category Detection.
	\label{tab:acd_mams}
	}
\end{table}

%% file: tables/zero_shot.tex
\begin{table}[t!]
\small
	\centering
	\setlength{\abovecaptionskip}{0.2cm}
    \setlength{\belowcaptionskip}{-0.6cm}
	\begin{tabular}{l|c|c}
		\toprule
		Model & R $\rightarrow$ M & M $\rightarrow$ R  \\
		\midrule
		BERT classification & 43.38  &  62.28  \\
		BART classification & 46.61 & 68.55  \\
		BART MLM & 47.86 & 70.64  \\
		BART generation & \textbf{49.84} & \textbf{72.46}  \\
		\bottomrule
	\end{tabular}
	\caption{Zero-Shot results: ACSA. R $\rightarrow$ M indicates training on Rest14 and testing on MAMS. M $\rightarrow$ R indicates training on MAMS and testing on Rest14.
	\label{tab:zero_shot}}
\end{table}

%% file: analysis.tex
\section{Analysis}

\input{figures/analysis}
\subsection{Influence of Category Frequency}
Aspect categories can be implicit and do not necessarily occur as terms in the given sentence. To explore the correlation between ACSA accuracy and the occurrence frequency of a given category, we split the eight categories in the MAMS test set into four subsets based on the occurrence frequency. The category (i.e., \emph{miscellaneous}) that never occurs in the given sentence is put into the {\it zero frequency} subset, the 15\% least frequent (i.e., \emph{ambience}, \emph{staff}) are put into {\it low frequency} subset, the 30\% most frequent (i.e., \emph{menu}, \emph{service}) are put into {\it high frequency} subset, and the remaining (i.e., \emph{price}, \emph{food}, \emph{place}) are put into {\it mid frequency} subset.

Figure~\ref{fig:occur} shows the accuracy of \textit{BART classification} and our model against the frequency. As the category occurrence frequency decreases, the relative gap of accuracy between the two models increases. In the {\it zero frequency}, our method gives absolutely 8.03\% stronger accuracy than \textit{BART classification}. This demonstrates that our method is more robust in summarizing the sentiment polarity of abstract or rare categories. Even if there are no explicit category terms in the sentence, the generation method can give the implicit category opinion of the whole sentence according to the context.
\input{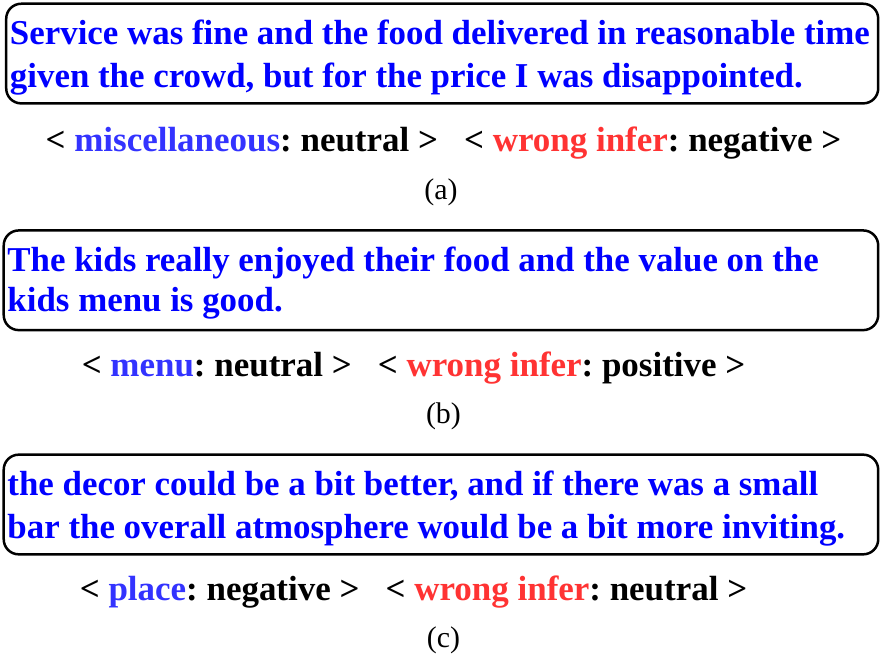}
\subsection{Case Study}
Figure \ref{fig:case} shows typical examples from the test set which cannot be inferred by the \textit{BART classification} model. In sentence (a), the given category {\it miscellaneous} does not occur as a term in the given sentence. Our method can synthesize different sentiment polarities with different aspects to obtain correct polarity. In sentence (b), “{\it the value on the kids menu is good}”, {\it good} modifies {\it the value}, rather than the given category {\it menu}. Our method gives the correct polarity, not being affected by the surrounding other aspect sentiments. The last instance (c) has conditional reasoning which is difficult for \textit{BART classification}. In contrast, {\it BART generation} gives the correct label by correctly recognizing the negativity in ``{\it if there was ... would be a bit more inviting}''. This is likely because our method makes use of pre-trained knowledge to infer the inter-sentential correlations between the input and the output sequences, which the \textit{BART classification} model failed to achieve due to the indirect use of BART in the additional classification network.

\section{Conclusion}
We investigated a generation method for aspect category detection (ACD) and aspect category sentiment analysis (ACSA), which can make better use of BART's advantages in making semantic level summaries to the input by not introducing additional model parameters. Experiments show that our proposed method obtains superior performance over the baseline models for both sentence-level and document-level aspect sentiment analysis. In contrast to the traditional sentiment classification methods, our method is also more powerful on zero-shot and few-shot tasks. 

%% file: figures/analysis.tex
\begin{figure}
	\centering
	\setlength{\abovecaptionskip}{0cm}
    \setlength{\belowcaptionskip}{-0.6cm}	
    \includegraphics[scale=0.3]{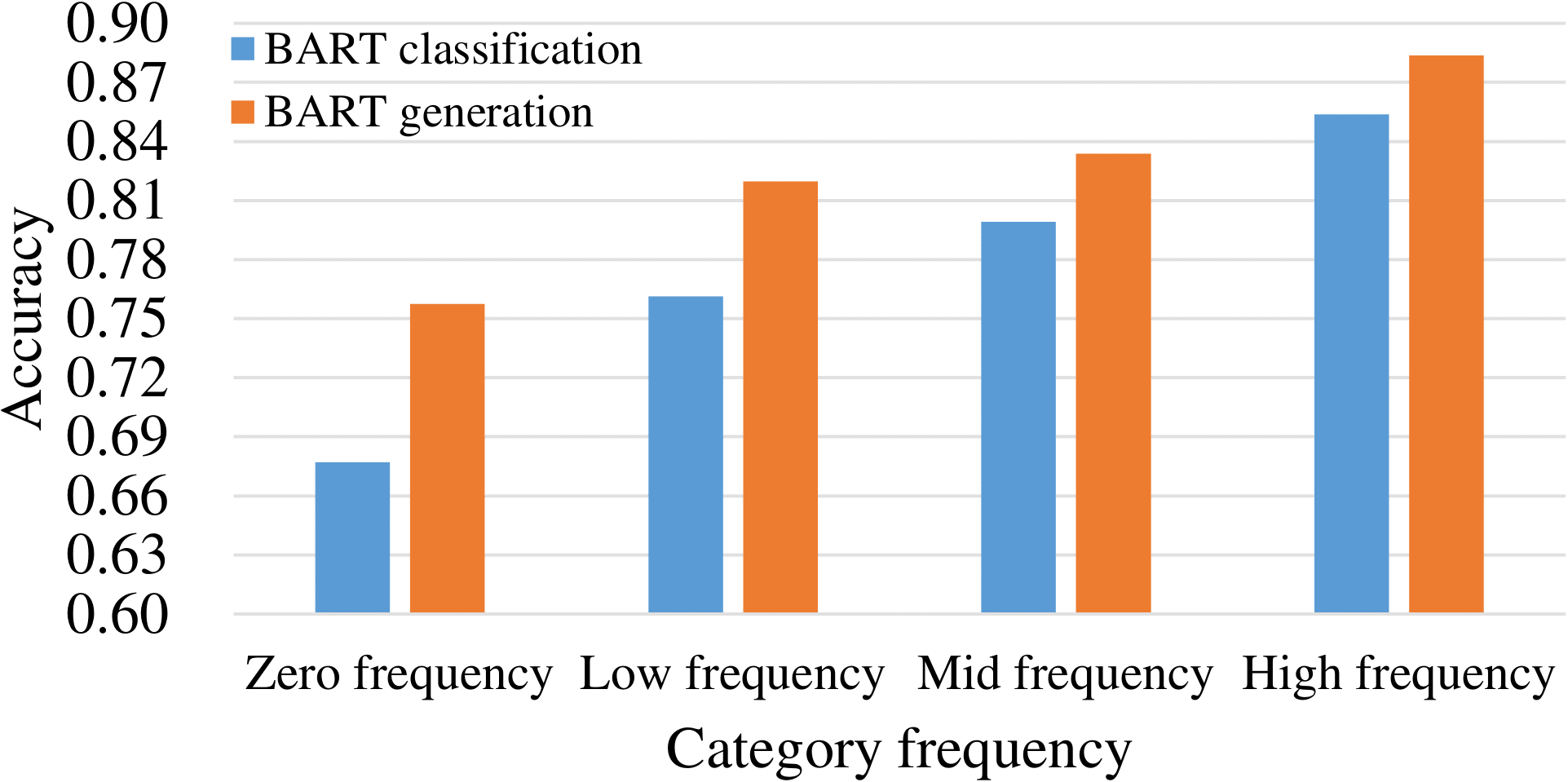}
	\caption{Comparison of accuracy with different category frequency on MAMS.}
	\label{fig:occur}
\end{figure}

%% file: figures/case.tex
\begin{figure}
	\centering
	\setlength{\abovecaptionskip}{0cm}
    \setlength{\belowcaptionskip}{-0.63cm}
	\includegraphics[scale=0.7]{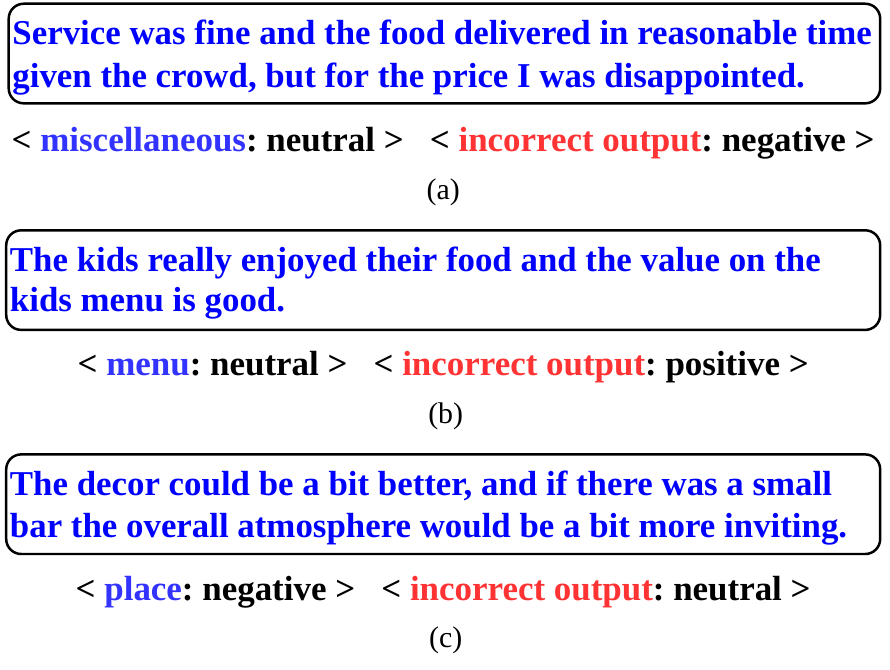}
	\caption{Examples of \textit{BART classification}. (a) is an instance with category do not occur as term in sentence. (b) represents that our method is not affected by the surrounding interference information. (c) needs conditional reasoning for analysis. Our method can obtain
    correct sentiment polarity.}
	\label{fig:case}
\end{figure}

%% file: appendix.tex
\appendix

\section{Datasets}
\label{sec:appendix}

\subsection{Sentence-Level Datasets}
\paragraph{Rest14} \citep{pontiki-etal-2014-semeval} Following previous work \citep{cheng2017aspect,tay2018learning,hu2019can}, we remove samples with conflict polarities. Since there is no official development set for Rest14, we use the split offered by \citet{tay2018learning}.

\paragraph{Rest14-hard} Following \citet{xue2018aspect}, we construct Rest14-hard, where the training set and development set are the same as Rest14’s, while test set  is constructed from the test set of Rest14. The test set of Rest14-hard only includes sentences containing at least two aspect categories with different sentiment polarities.
\paragraph{MAMS}\citet{jiang2019challenge} Since the test set of Rest14-hard is small, we also adopt the Multi-Aspect Multi-Sentiment dataset for Aspect Category Sentiment Analysis (denoted by MAMS). All sentences in MAMS contain multiple aspect categories with different sentiment polarities.

\subsection{Document-Level Datasets}
TripAdvisor \citep{WangLZ10} and BeerAdvocate \citep{McAuley,Lei} contain seven aspects (value, room, location, cleanliness, check in/front desk, service, and business service) and four aspects (feel, look, smell, and taste) respectively. We randomly split them into training, development, and testing sets with 80/10/10\%.
~\\

Statistics of these three sentence-level datasets are given in Table~\ref{tab:dataset1} and two document-level datasets are described in Table \ref{tab:dataset2}.
\input{tables/dataset1}
\input{tables/dataset2}

\section{Settings}
Each method is trained for 30 epochs, during which the model with the best performance on the validation set is saved. We also apply early stopping in training, which means that the training will stop if the performance on validation set does not improve in 5 epochs.



%% file: tables/dataset1.tex
\begin{table}
	\centering
	\setlength{\abovecaptionskip}{0.1cm}
    \setlength{\belowcaptionskip}{-0.2cm}
	\footnotesize
	\begin{tabular}{ c|c|c|c|c } 
		\toprule
		\multicolumn{2}{c|}{Dataset} & Pos. & Neg. & Neu. \\ 
		\midrule
		\multirow{3}*{Rest14} & Train & 1855 & 733 & 430 \\ 
		\cmidrule{2-5}
		& Dev & 324 & 106 & 70 \\ 
		\cmidrule{2-5}
		& Test & 657 & 222 & 94 \\ 
		\midrule
		Rest14-hard & Test & 21 & 20 & 12 \\ 
		\midrule
		\multirow{3}*{MAMS-ACSA}  & Train & 1929 & 2084 & 3077 \\ 
		\cmidrule{2-5}
		& Dev & 241 & 259 & 388 \\ 
		\cmidrule{2-5}
		& Test & 245 & 263 & 393 \\ 
		\bottomrule
	\end{tabular}
	\caption{Statistics of the sentence-level datasets.}
	\label{tab:dataset1}
\end{table}

%% file: tables/dataset2.tex
\begin{table}[t!]
	\centering
	\setlength{\abovecaptionskip}{0.1cm}
    \setlength{\belowcaptionskip}{-0.6cm}
	\footnotesize
	\begin{tabular}{l|c|c|c}
		\toprule
		Dataset & \#docs & \#words/doc & words/sent  \\
		\midrule
		TripAdvisor & 29,391  & 251.7  & 18.0 \\
		BeerAdvocate & 51,020  & 144.5  &  12.1 \\
		\bottomrule
	\end{tabular}
	\caption{Statistics of the document-level datasets. The rating scale of TripAdvisor dataset is 1-5. The rating scale of BeerAdvocate dataset is 1-10.
	\label{tab:dataset2}
	}
\end{table}